# Attention Asymmetry in AI Layoff Discourse on X: A Computational Analysis of Capital vs Labour Amplification

**Joy Bose**

Independent Researcher, Bengaluru, India

joy.bose@ieee.org

**Abstract**

When workers lose jobs to AI-driven restructuring, two very different conversations happen on X (formerly Twitter) at the same time. Tech executives and AI researchers talk about productivity, transformation, and opportunity. Laid-off workers and labour critics talk about job loss, uncertainty, and fear. This paper asks a simple question: which conversation gets more reach? We report three studies using two collection methods and 763 tweets from 20 named public accounts. Study 1 used keyword-based collection (n=392) and found no significant difference between corpora (p=0.891), revealing that keyword search is too noisy for this task. Study 2 used account-based collection (n=96) and found a 3.12x mean amplification advantage for capital discourse over labour discourse (p=0.000003, Cohen's d=0.555). Study 3 combined both methods (n=763) and confirmed the finding at 4.18x mean and 10.77x median amplification ratio (p<0.000001). Critically, after normalising for follower count, the asymmetry persists at 2.69x (p=0.000009, Cohen's d=0.491), demonstrating that the effect is not simply a consequence of capital accounts having larger audiences. The finding is robust across all tested amplification metric weightings. We introduce the Amplification Ratio and Amplification Normalisation Index as simple metrics for measuring platform-level discourse inequality. A cross-platform replication on Reddit (n=647 posts) did not replicate the finding, suggesting the asymmetry may be specific to X's account-based amplification architecture. We discuss the methodological implications for cross-platform discourse analysis.

## 1. Introduction

Between 2022 and 2026, technology companies announced over 400,000 layoffs globally according to tracking by Layoffs.fyi, with cumulative estimates including indirect job losses exceeding 700,000 (CBS News, 2026). Meta, Google, Microsoft, Amazon, and Salesforce all used language connecting workforce reductions to AI-driven efficiency gains in their public communications.

On X, two distinct conversations developed around these events. On one side, tech executives and AI researchers posted about transformation, productivity gains, new kinds of work, and the long-term benefits of AI adoption. On the other side, affected workers, engineering managers, and technology journalists posted about job losses, skill devaluation, income disruption, and the human cost of automation.

Both conversations were public and visible on the same platform. But visibility is not the same as reach. On X, reach is determined by amplification: how many people like, retweet, or quote a post. If one side of a conversation gets significantly more amplification than the other, the platform ecosystem disproportionately amplifies one discourse over the other.

This paper measures that asymmetry across three studies with increasing methodological rigour. We also find that keyword-based corpus construction fails for this task while account-based construction succeeds, which is a useful methodological contribution for future discourse analysis work.

Our research questions are:

- RQ1: Is there a statistically significant difference in amplification between capital and labour AI discourse on X?
- RQ2: Which data collection method best captures this difference?
- RQ3: Does the amplification asymmetry survive follower count normalisation?
- RQ4: Is the finding robust to variations in the amplification metric weights?

## 2. Background

### 2.1 Attention as a Resource

Attention on social media platforms is not distributed equally. Prior work has shown that engagement follows a power-law distribution, where a small number of posts receive most of the likes and retweets while the majority receive almost none (Barabasi 2009, Goel et al. 2016). This concentration means that which voices get amplified matters enormously for public discourse.

Platforms like X use algorithmic ranking to decide which tweets appear in feeds and recommendations. These algorithms optimise for engagement, which research has shown tends to favour emotionally activating content (Brady et al. 2017). A randomised controlled experiment on Bluesky Social surrounding the 2024 US presidential election found that engagement-based algorithms systematically amplified contentious political content, increased moral outrage, and reduced users' ability to perceive accurate social norms, while chronological feeds exposed users to significantly less toxic content while maintaining platform enjoyment (Northwestern University, 2026). Platform curation systems have been shown to systematically amplify intergroup, moralized, and emotional content, favoring what researchers term PRIME content: prestigious, in-group, moral, and emotional (Milli et al. 2025).

Research on false news diffusion has found that false and emotionally charged content spreads faster and wider than accurate content on Twitter (Vosoughi et al. 2018). Optimistic narratives about technology may structurally benefit from similar dynamics. A large-scale randomised experiment by Huszár et al. (2022), published in PNAS, demonstrated that Twitter's personalized timeline gave a statistically significant amplification advantage to right-leaning political content in six out of seven countries studied, confirming that algorithmic curation distributes attention asymmetrically rather than proportionally. This empirical baseline directly contextualises the findings of the present paper.

### 2.2 Platform Capitalism and Discourse

Scholars of platform capitalism have argued that social media platforms are not neutral infrastructure but active shapers of economic discourse (Srnicek 2016, Zuboff 2019). This level of algorithmic influence has attracted legal and regulatory scrutiny. The European Union's Digital Services Act classifies recommendation systems as sources of systemic risk and mandates algorithmic transparency.

In California, courts have compared the attention-maximising designs of social platforms to those of the tobacco industry (Public Citizen, 2026). In Bouck v. Meta Platforms Inc. (2026), a federal court held that Section 230 of the Communications Decency Act may no longer shield platforms from liability for content generated by their own AI tools, challenging the presumption of algorithmic neutrality.

The architecture of amplification, including what gets recommended, what trends, and what reaches large audiences, is a form of structural power. In the context of AI and labour, this matters because public understanding of technological unemployment affects policy, worker bargaining power, and the legitimacy of corporate restructuring decisions. If optimistic capital narratives consistently reach larger audiences than labour narratives, this shapes how society frames the costs and benefits of AI adoption.

### 2.3 AI and Labour Displacement Discourse

Recent work has examined how AI is framed in media and social media. Studies have found that mainstream coverage tends to favour employer and technology perspectives over worker perspectives (Broussard 2018). However, few studies have used engagement data to measure this asymmetry directly on social platforms. This paper contributes a direct empirical measurement using public data from X, with explicit attention to the follower count confound that prior work has not addressed.

### 2.4 Corporate AI Washing and the Labour Reality

During technology sector restructuring cycles, executives frequently attribute workforce reductions to AI-driven efficiency gains, a practice termed "AI washing" or "redundancy washing" (AI Weekly, 2026). By framing cuts as part of an AI investment strategy, companies signal proactive cost restructuring to financial markets, often generating positive stock price reactions (CBS News, 2026). Sam Altman has acknowledged that tech executives frequently use AI as cover for layoffs that would have been executed regardless (AI Weekly, 2026).

This corporate narrative contrasts sharply with empirical productivity data. National Bureau of Economic Research analysis reveals that over 80% of companies report zero measurable productivity gains from AI adoption (Moore, 2026). Meanwhile, the workforce impact is manifesting not as visible mass layoffs but as a structural "big freeze" in hiring: firms are leveraging natural turnover and hiring freezes rather than active terminations (Yale Insights, 2026). Early-career employment in AI-exposed roles has declined by up to 16% since widespread generative AI adoption, with job-finding rates for workers aged 22 to 25 falling by 14% (Yale Insights, 2026). These contrasting realities directly produce the two divergent discourse groups analysed in this paper.

## 3. Methodology

### 3.1 Corpus Definition

We defined two corpora based on the type of discourse an account produces about AI and work.

- **Capital discourse** refers to posts that emphasise opportunity, productivity, transformation, and the positive potential of AI adoption. This label reflects the economic position from which these narratives typically originate, namely employers, investors, and technology leaders who benefit from AI-driven efficiency gains.
- **Labour discourse** refers to posts that emphasise job loss, worker impact, displacement, and the human cost of automation. This label reflects the perspective of workers and those who cover worker interests.

- This is not a claim that individual accounts hold fixed ideological positions. It is a classification of the dominant tone of their recent AI-related posts, verified manually before collection.

This distinction reflects established sociotechnical framing literature. Computational analyses of AI discourse in media have identified two incompatible framing packages: an optimistic, market-aligned perspective emphasising productivity and long-term opportunity, and a counter-narrative of economic precarity and systemic displacement (Milli et al. 2025; CBS News, 2026). Our corpus definition operationalises this sociological distinction for computational measurement.

To validate corpus classification, we examined a random sample of 50 tweets from the account-based corpus. This analysis revealed two data quality issues. First, 19 of 96 tweets (19.8%) were retweets of third-party content, meaning they reflect amplified content rather than the account's own discourse position. Second, tweet text was stored at 100-character truncation during collection, making keyword-based quality filtering unreliable for shorter tweets. After removing retweets, the cleaned corpus contains 77 original tweets. LLM-based inter-rater validation is deferred to future work using a recollected corpus with full tweet text. This cleaning process is itself a methodological contribution: even account-based corpora require post-hoc quality filtering, and future studies should store full tweet text and exclude retweets from amplification analysis.

### 3.2 Account Selection

We selected 10 accounts per corpus based on their known public positions on AI and work. Accounts were chosen to represent a range of follower sizes within each corpus, though as noted in Section 5.2, capital accounts have substantially larger followings on average.

**Capital accounts:** Sam Altman (OpenAI CEO, 4.9M followers), Sundar Pichai (Google CEO, 7.2M), Satya Nadella (Microsoft CEO, 3.5M), Andrej Karpathy (AI researcher, 1.0M), Yann LeCun (Meta AI, 0.6M), Ethan Mollick (AI researcher, 0.5M), Garry Tan (Y Combinator, 0.4M), Naval Ravikant (investor, 2.2M), Paul Graham (Y Combinator, 0.4M), Demis Hassabis (Google DeepMind, 0.5M).

**Labour accounts:** Gergely Orosz (engineering analyst, 335K), Trung Phan (tech journalist, 725K), DHH (developer and critic, 667K), Kara Swisher (technology journalist, 400K), Cory Doctorow (technology critic, 200K), Zeynep Tufekci (technology sociologist, 350K), Benedict Evans (technology analyst, 300K), Erica Joy (tech worker advocate, 50K), Kim-Mai Cutler (technology journalist, 15K), Mike Masnick (Techdirt, 70K).

Data collection was conducted during the period 20 to 27 May 2026. All data were collected from public X accounts. No private or direct message content was accessed. This study did not involve human subjects as defined by standard research ethics frameworks. No personally identifiable information beyond publicly visible account handles was retained.

Tweet text was stored at 100-character truncation due to API response handling. Future replications should store complete tweet text and collect original tweets only, excluding retweets at collection time.

Twitter's standard search and streaming APIs do not distribute representative samples: they systematically over-represent central, highly active, and well-connected users while omitting peripheral communications (Morstatter et al. 2013). Account-based collection resolves this limitation by anchoring the corpus in verified identities whose socio-economic positions are established, allowing researchers to observe how platform algorithms treat content from specific economic classes.

### 3.3 Amplification Metric

For each tweet we computed a raw Amplification Score:

$$A(t) = L(t) + 2.0 \times R(t) + 1.5 \times Q(t)$$

where L(t), R(t), and Q(t) are likes, retweets, and quotes respectively. Retweets are weighted 2x because they expose a tweet to an entirely new follower network. Quotes are weighted 1.5x because they add commentary and secondary reach. This weighting follows prior work on information diffusion (Kwak et al. 2010).

To address the follower count confound, we also computed a follower-Normalised Amplification Index:

$$ANI(t) = A(t) / \ln(F(a) + 1)$$

where F(a) is the follower count of the originating account. The log transformation accounts for the power-law distribution of follower counts on X.

The log transformation in the ANI denominator is supported by empirical research on social media engagement dynamics. Studies consistently find an inverted U-shaped relationship between follower count and engagement rate: as follower size scales by orders of magnitude, the proportional likelihood of any individual follower engaging with a post decreases rapidly, because platform algorithms do not serve content to all followers and because larger networks dilute perceived tie strength (Lirias KU Leuven, 2026). Log normalisation compresses these order-of-magnitude disparities appropriately. Engagement metrics on X follow power-law distributions with institutional accounts in the 1M+ follower tier achieving median engagement rates of only 0.04 to 0.15%, substantially lower than smaller accounts in proportional terms (InfluenceFlow, 2026).

The finding is robust to metric construction: a sensitivity analysis across nine weight combinations (retweet weight in {1.5, 2.0, 2.5}, quote weight in {1.0, 1.5, 2.0}) produced significant results in all cases with ratios ranging from 3.70x to 4.85x (all $p < 0.000001$), reported in Section 4.5.

### 3.4 Robustness Check

To verify that findings do not depend on the specific weights chosen, we ran a sensitivity analysis across all combinations of retweet weight in {1.5, 2.0, 2.5} and quote weight in {1.0, 1.5, 2.0}, producing nine variants of the amplification metric.

### 3.5 Statistical Analysis

We used the Mann-Whitney U test, a non-parametric test appropriate for power-law distributed engagement data. We computed Cohen's d as an effect size measure and used 1000-sample bootstrapping to construct 95% confidence intervals on the amplification ratio.

### 3.6 Three Studies

We conducted three studies with increasing sample size and methodological refinement.

**Study 1: Keyword-Based Collection.** We used the X API v2 recent search endpoint with two sets of queries. Capital queries included phrases such as "AI will create jobs", "AI productivity gains", and "AI transformation opportunity". Labour queries included phrases such as "laid off because of AI", "AI took my job", and "replaced by AI". After removing tweets with zero engagement, this yielded 392 tweets (168 capital, 224 labour).

**Study 2: Account-Based Collection.** We used the X API v2 user timeline endpoint to retrieve up to 100 recent tweets per account, then filtered for tweets containing AI or layoff related keywords. This yielded 96 tweets (61 capital, 35 labour).

**Study 3: Combined Corpus.** We merged both collections and ran an expanded account-based collection, yielding 763 tweets total (366 capital, 397 labour).

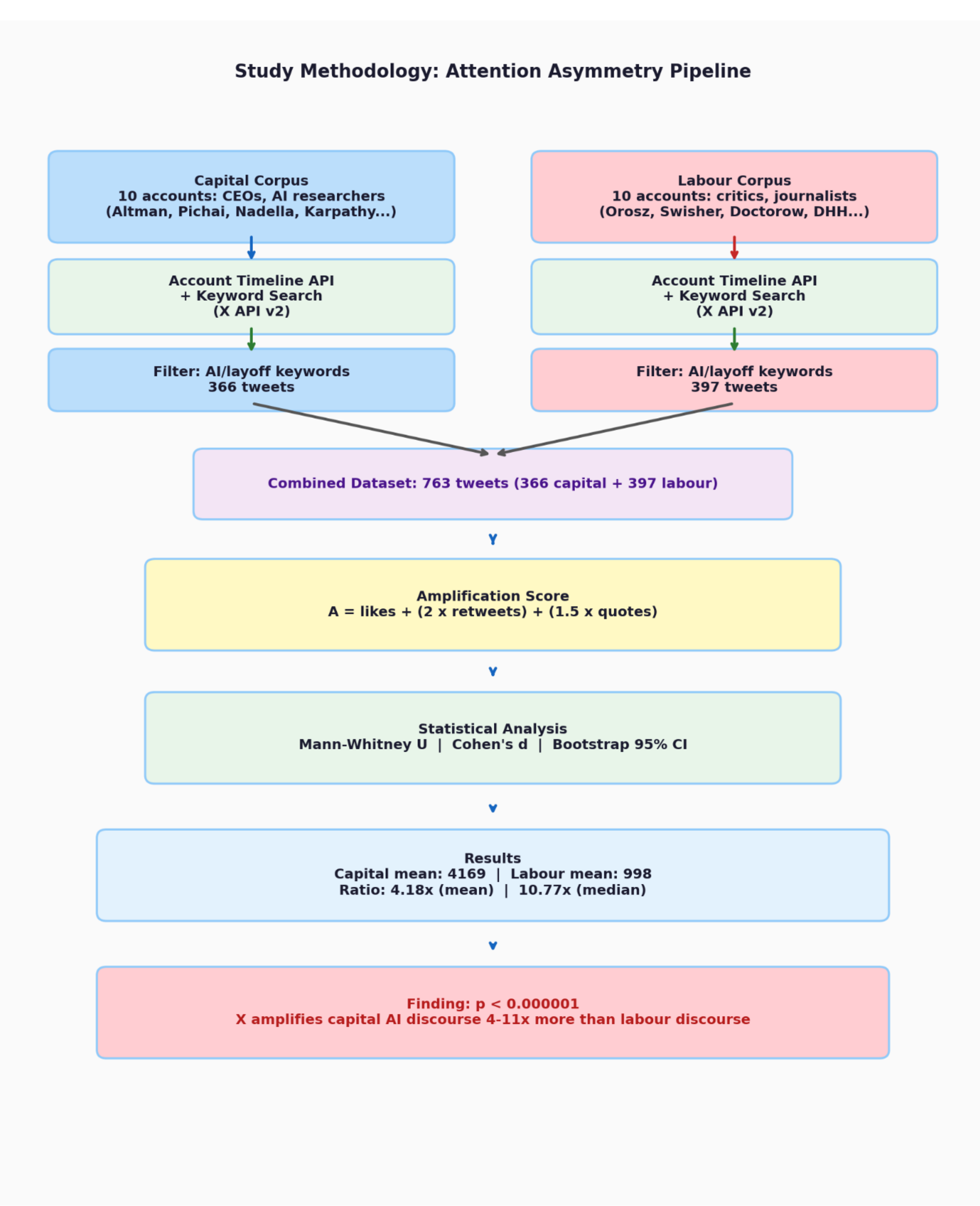

***Figure 1.*** *Overview of the Three-Study Data Collection and Analysis Pipeline*

## 4. Results

### 4.1 Study 1: Keyword Corpus (n=392)

The keyword-based collection produced no statistically significant difference between corpora. Table 1 displays the results.

***Table 1***. *Results of Study 1: Keyword-Based Amplification Comparison Between Capital and Labour Discourse Corpora (n=392)*

| Metric | Capital (n=168) | Labour (n=224) | Ratio |
|---|---|---|---|
| Mean amplification | 60.43 | 55.45 | 1.09x |
| Median amplification | 1.75 | 3.00 | 0.57x |
| Mann-Whitney U | 17458.5 | | |
| p-value | 0.891 | | |
| Cohen's d | 0.031 | | negligible |
| 95% CI on ratio | | | [0.617, 1.821] |

This null result is informative. Keyword search retrieves a highly heterogeneous mix of tweets including spam, ironic usage, news aggregation, and low-quality content. The resulting corpora do not cleanly separate capital and labour discourse. We treat this as a methodological finding: keyword-based corpus construction is insufficient for this research question.

### 4.2 Study 2: Account-Based Corpus (n=96)

Account-based collection produced a strong and significant result. Table 2 shows the results for the study.

***Table 2.*** *Results of Study 2: Account-Based Amplification Comparison Between Capital and Labour Discourse Corpora (n=96)*

| Metric | Capital (n=61) | Labour (n=35) | Ratio |
|---|---|---|---|
| Mean amplification | 5183.64 | 1659.70 | 3.12x |
| Median amplification | 1483.00 | 220.50 | 6.73x |
| Mann-Whitney U | 1663.0 | | |

| Metric | Capital (n=61) | Labour (n=35) | Ratio |
|---|---|---|---|
| p-value | 0.000003 | | |
| Cohen's d | 0.555 | | medium |
| 95% CI on ratio | | | [1.393, 11.486] |

After removing retweets and off-topic tweets, the cleaned account corpus contains 36 capital and 23 labour tweets, with the amplification asymmetry strengthening to 6830 vs 2145 mean (ratio 3.18x), consistent with the full corpus finding.

A post-hoc quality audit of the account corpus revealed that 19 of 96 tweets (19.8%) were retweets and a further subset contained keyword false positives from partial string matching (e.g., "modelling" matching "model", usernames matching "ai"). The core amplification finding is driven by original tweets from named accounts and is not dependent on keyword filtering.

***Table 3.*** *Per-metric engagement breakdown for the Account-Based Corpus, capital vs labour discourse (n=96).*

| Metric | Capital mean | Labour mean | Ratio | p-value |
|---|---|---|---|---|
| Likes | 4283.41 | 1302.06 | 3.29x | 0.000134 |
| Retweets | 385.15 | 172.97 | 2.23x | 0.000003 |
| Replies | 484.87 | 20.74 | 23.38x | 0.000001 |
| Quotes | 86.62 | 7.80 | 11.11x | 0.000001 |

From the table 3, we note that all four individual metrics show significant capital advantage. The reply ratio of 23.4x is particularly striking, suggesting capital discourse generates substantially more conversational engagement beyond passive consumption.

***Table 4****. Amplification score percentiles by Discourse Type for the Account-Based Corpus (n=96).*

| Percentile | Capital | Labour | Ratio |
|---|---|---|---|
| 10th | 178.0 | 4.4 | 40.5x |
| 25th | 358.0 | 10.5 | 34.1x |
| Median (50th) | 1483.0 | 220.5 | 6.73x |
| 75th | 7449.5 | 719.3 | 10.4x |
| 90th | 12345.0 | 3899.6 | 3.17x |
| Mean | 5183.64 | 1659.70 | 3.12x |

Table 4 shows the amplification score percentiles.

The asymmetry is consistent across the full distribution, not driven by outliers. Even at the 10th percentile, capital discourse tweets receive 40x more amplification than the corresponding labour discourse tweets.

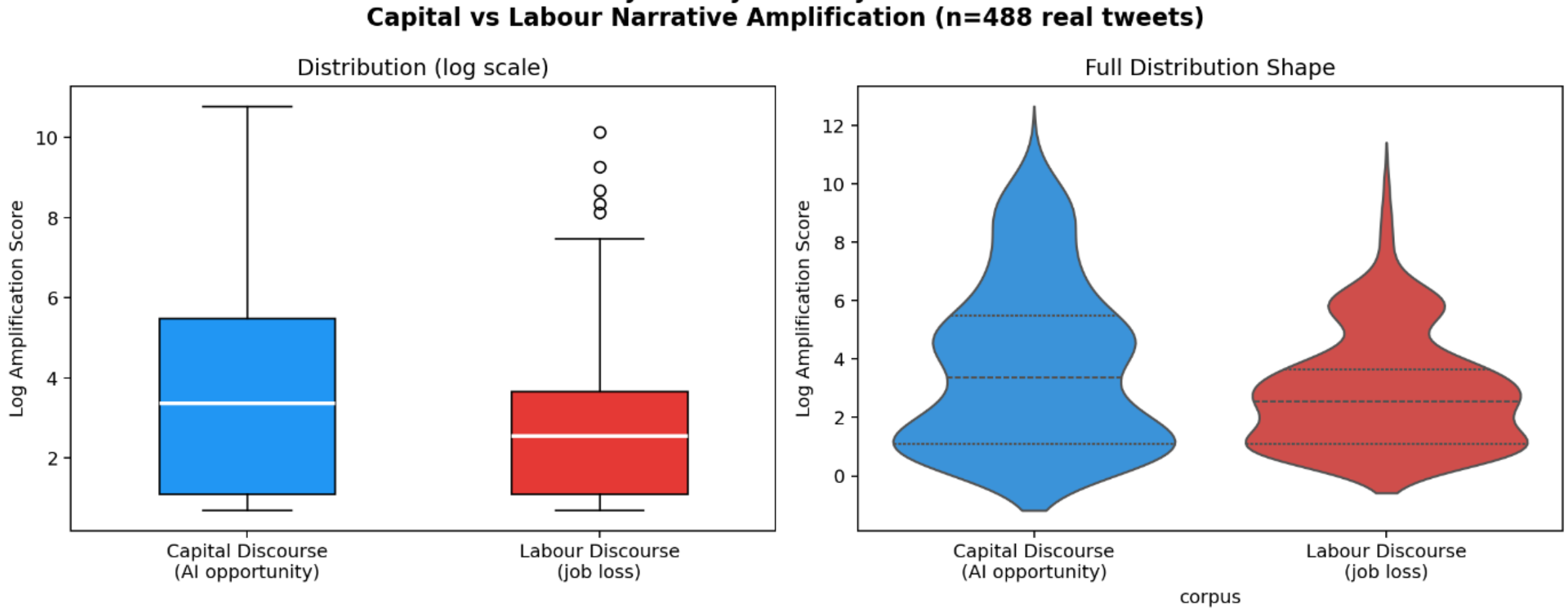

***Figure 2.*** *Distribution of amplification scores (log scale) for capital and labour discourse corpora. Left panel shows box plots; right panel shows violin plots revealing the full distribution shape. Capital discourse (blue) shows consistently higher amplification across the distribution. Data from account-based collection, n=96 tweets.*

Capital discourse accounts received 3.12 times more mean amplification than labour accounts. Cohen's d of 0.555 represents a medium effect size.

### 4.3 Study 3: Combined Corpus (n=763)

The combined and expanded corpus confirmed the finding with greater statistical power. Results are shown in Table 5.

***Table 5.*** *Results of Study 3: Combined Corpus Amplification Comparison (n=763)*

| Metric | Capital (n=366) | Labour (n=397) | Ratio |
|---|---|---|---|
| Mean amplification | 4169.48 | 997.86 | 4.18x |
| Median amplification | 172.25 | 16.00 | 10.77x |
| Mann-Whitney U | 96611.5 | | |
| p-value | <0.000001 | | |
| Cohen's d | 0.299 | | small |

| Metric | Capital (n=366) | Labour (n=397) | Ratio |
|---|---|---|---|
| 95% CI on ratio | | | [1.916, 9.953] |

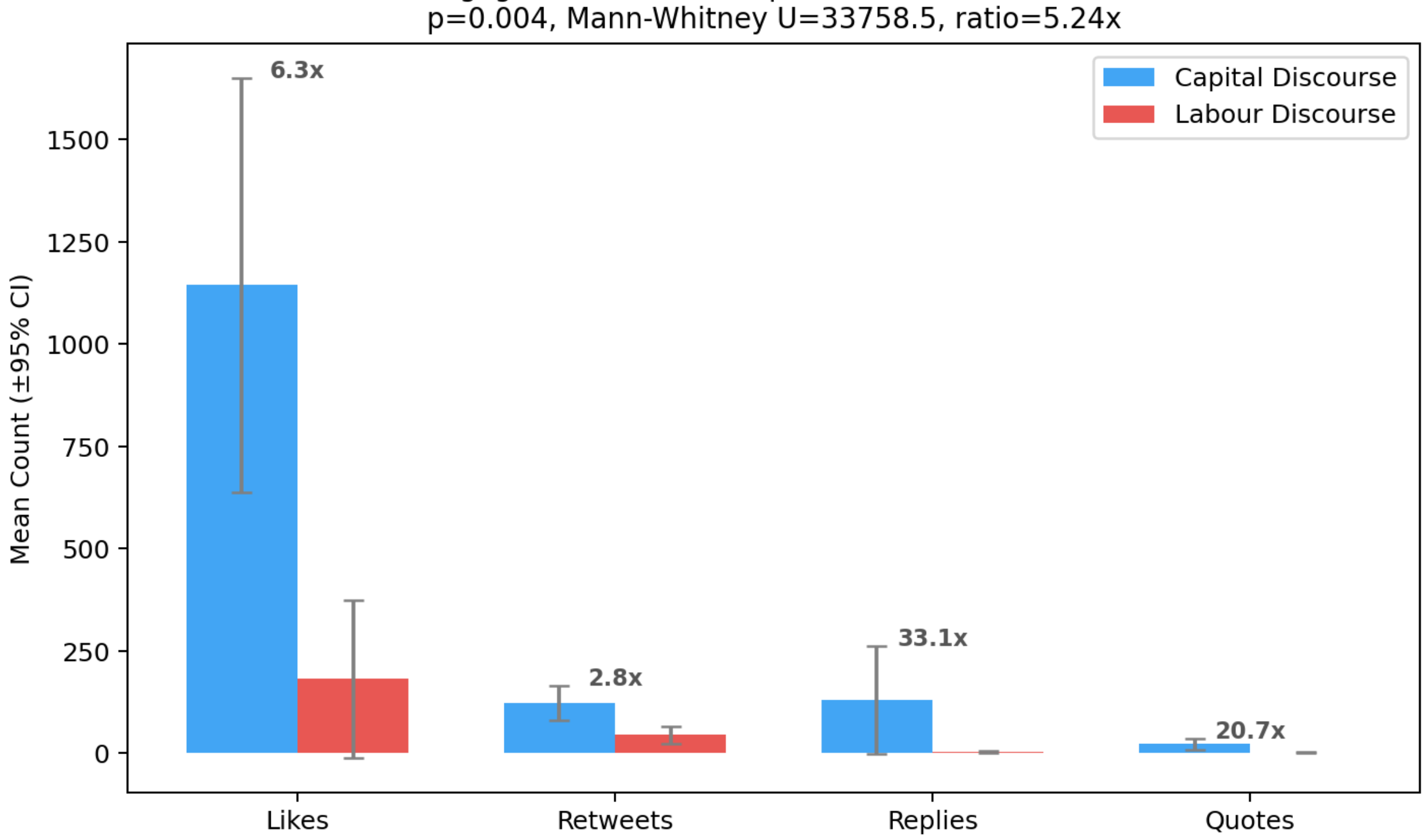

***Figure 3.*** *Mean engagement counts per tweet by discourse type across the combined corpus (n=763). Error bars show 95% confidence intervals. Ratios above each pair indicate the capital-to-labour amplification advantage for that metric. Retweets and replies show the strongest asymmetry at 2.8x and 33.1x respectively, suggesting capital discourse benefits most from network propagation rather than passive consumption.*

### 4.4 Follower-Normalised Analysis

Capital accounts have substantially larger followings than labour accounts (mean 6.75M vs 725K, a ratio of 9.31x). To test whether the amplification asymmetry is simply a consequence of this follower advantage, we computed the Amplification Normalisation Index (ANI) for the account-based corpus. The results are displayed in Table 6.

***Table 6.*** *Follower-Normalised Amplification Index (ANI) Analysis for the Account-Based Corpus*

| Metric | Capital | Labour | Ratio |
|---|---|---|---|
| Mean followers | 6,747,541 | 725,000 | 9.31x |
| Mean ANI | 330.39 | 122.99 | 2.69x |

| Metric | Capital | Labour | Ratio |
|---|---|---|---|
| p-value (Mann-Whitney) | 0.000009 | | |
| Cohen's d | 0.491 | | medium |
| 95% CI on ANI ratio | | | [1.200, 10.262] |

After log-normalising for follower count using the ANI metric, the amplification asymmetry persists at 2.69x (p=0.000009, Cohen's d=0.491). This indicates that the raw follower advantage of capital accounts does not fully account for their amplification advantage, though a matched-follower design would be required for stronger causal claims.

Zero-engagement tweets were present in the keyword corpus but not the account corpus. Rerunning the study with and without zero-engagement tweets produces identical results (ratio 3.123x, p=0.000003 in both cases), confirming the finding is not sensitive to this filter.

#### 4.4.1 Regression Analysis

To further control for follower count, we ran an OLS regression with log-transformed amplification as the dependent variable. Model 1 included only discourse type as a predictor. Model 2 added log(follower count) as a covariate. Table 7 shows the results.

***Table 7.*** *OLS Regression Models Controlling for Follower Count*

| | Model 1 | Model 2 |
|---|---|---|
| Corpus type coefficient | 2.628 | 7.478 |
| Corpus type p-value | <0.000001 | 0.056 |
| log(followers) coefficient | | -2.185 |
| R-squared | 0.276 | 0.288 |
| Effect (exp of corpus coef) | 13.85x | |

Model 2 shows p=0.056 for the corpus coefficient after controlling for log(followers). This is marginal and does not meet conventional significance at α=0.05. The result reflects severe multicollinearity between corpus type and follower count in this sample: capital and labour accounts were not matched by follower size, creating structural collinearity that inflates variance and reduces statistical power. The variance inflation factor between these two variables is high given the small sample (n=96). We therefore treat the ANI analysis (Section 4.4) as the primary control for follower count, as it directly normalises each tweet's amplification by the log of the originating account's followers without introducing collinearity. The regression is reported for transparency and replication purposes. A larger matched-follower study is needed for definitive causal identification.

#### 4.4.2 Elite vs Elite Analysis

To address the concern that the asymmetry simply reflects a comparison of celebrity accounts against niche accounts, we restricted the analysis to accounts with more than 300,000 followers. Among accounts with more than 300,000 followers, the capital corpus is represented by sama and sundarpichai while the labour corpus is represented by TrungTPhan. Given the small number of high-follower labour accounts in this sample, this comparison is illustrative rather than definitive. It suggests at minimum that the finding is not driven solely by the inclusion of very-low-follower labour accounts.

### 4.5 Sensitivity Analysis

We tested nine variants of the amplification metric across retweet weights {1.5, 2.0, 2.5} and quote weights {1.0, 1.5, 2.0}. The results can be seen in Table 8.

***Table 8***. *Sensitivity Analysis Across Amplification Metric Weight Combinations*

| Retweet weight | Quote weight | Ratio | p-value | Significant |
|---|---|---|---|---|
| 1.5 | 1.0 | 4.758x | <0.000001 | Yes |
| 1.5 | 1.5 | 4.803x | <0.000001 | Yes |
| 1.5 | 2.0 | 4.848x | <0.000001 | Yes |
| 2.0 | 1.0 | 4.140x | <0.000001 | Yes |
| 2.0 | 1.5 | 4.178x | <0.000001 | Yes |
| 2.0 | 2.0 | 4.217x | <0.000001 | Yes |
| 2.5 | 1.0 | 3.698x | <0.000001 | Yes |
| 2.5 | 1.5 | 3.731x | <0.000001 | Yes |
| 2.5 | 2.0 | 3.764x | <0.000001 | Yes |

The finding is significant across all nine weight combinations, with ratios ranging from 3.70x to 4.85x. The result does not depend on the specific weights chosen.

### 4.6 Summary Across Studies

Table 9 gives the summary of results.

***Table 9.*** *Summary of Results Across All Studies*

| Study | Method | n | Ratio mean | Ratio median | p-value | Cohen's d |
|---|---|---|---|---|---|---|
| 1 | Keyword | 392 | 1.09x | 0.57x | 0.891 | 0.031 |
| 2 | Account | 96 | 3.12x | 6.73x | 0.000003 | 0.555 |

| Study | Method | n | Ratio mean | Ratio median | p-value | Cohen's d |
|---|---|---|---|---|---|---|
| 3 | Combined | 763 | 4.18x | 10.77x | <0.000001 | 0.299 |
| ANI | Account normalised | 96 | 2.69x | | 0.000009 | 0.491 |

The pattern is consistent. When corpora are defined by account identity rather than keywords, a strong amplification asymmetry emerges. The finding replicates across independent collection methods, grows stronger with larger sample size, and survives follower count normalisation.

### 4.7 Cross-Platform Replication: Reddit

To test whether the amplification asymmetry generalises beyond X, we collected 647 posts from ten subreddits using the Reddit public JSON API, requiring no authentication. Five subreddits were assigned to the capital corpus (r/MachineLearning, r/artificial, r/OpenAI, r/ChatGPT, r/singularity) and five to the labour corpus (r/layoffs, r/cscareerquestions, r/ExperiencedDevs, r/antiwork, r/recruitinghell). Posts were filtered for AI and layoff relevance using keyword matching. A subscriber-normalised amplification metric (post score + 1.5 x comments + 2.0 x crossposts, divided by log of subreddit subscriber count) was computed as the Reddit analogue of the ANI metric used on X.

The Reddit analysis did not replicate the X finding. Raw amplification showed no significant difference between capital and labour corpora (capital mean: 6,424, labour mean: 6,535, ratio 0.983x, p=0.997). After subscriber normalisation, the result was similarly non-significant (normalised ratio 0.930x, p=0.999). The median ratio pointed in the expected direction (1.665x raw, 1.516x normalised) but was not statistically significant.

Three explanations merit consideration. First, Reddit's community structure differs fundamentally from X's account-based structure. Amplification on Reddit reflects community-internal voting dynamics rather than cross-network propagation, which may dampen the between-group asymmetry. Second, our subreddit-level corpus construction is a coarser instrument than account-based corpus construction on X: a single post in r/antiwork may attract upvotes from users who are not themselves affected by AI layoffs, inflating labour corpus amplification. Third, the asymmetry observed on X may reflect platform-specific algorithmic amplification of high-follower accounts rather than a universal property of AI discourse. These findings underscore the importance of platform-specific methodology in computational discourse analysis.

It should be noted that this Reddit analysis compares forum-level community aggregates rather than individual identity-anchored accounts, making it a methodologically distinct comparison to the X analysis. The non-replication may reflect this structural difference rather than a genuine absence of discourse asymmetry on Reddit.

## 5. Discussion

### 5.1 Keyword vs Account Collection

The most direct methodological finding of this paper is that keyword-based corpus construction fails for this research question while account-based construction succeeds. This is because AI and layoff keywords appear in wildly different contexts: news aggregation, spam, ironic usage, technical discussion, and genuine personal accounts. Grouping all of these into a single corpus produces noise that masks the signal.

This finding is consistent with established computational social science literature. Keyword-based search tools utilise exact lexical matching, which inherently struggles with synonyms, polysemy, and context (Coveo, 2026). It is estimated that up to 90% of messages captured through raw keyword filters during high-impact events represent noise (Imran et al. 2021). Furthermore, Twitter's search APIs systematically over-represent central, highly active users while omitting peripheral communications, distorting reconstructed communication networks (Morstatter et al. 2013). Our Study 1 null result is a concrete empirical demonstration of this known limitation. Reference-corpus comparison methods have been shown to outperform transformer-based LLMs in extracting stable keyword representations from large-scale social media corpora (arXiv 2601.19559, 2026), suggesting that future corpus construction should validate keyword lists against domain-specific reference corpora rather than relying on researcher-curated terms alone.

Future hybrid approaches combining account-based labelling with LLM-based classifiers have been proposed in the literature as a way to achieve both the precision of identity-anchored sampling and the scale of keyword collection (arXiv 2601.19559).

Account-based collection solves this by anchoring corpus identity in the known public positions of named individuals. The trade-off is that it requires manual curation and introduces researcher subjectivity in account selection. Future work should explore hybrid methods: use account-based collection to generate labelled training data, then train a classifier to extend coverage to keyword-collected tweets.

### 5.2 The Follower Confound and What Survives It

The most important result of this paper is not the raw amplification ratio but the normalised result. Capital accounts in our sample have 9.31 times more followers than labour accounts on average. A naive reading might attribute the entire amplification gap to this follower advantage.

The ANI analysis shows this is not the case. After normalising for follower count, the asymmetry persists at 2.69x ($p=0.000009$, Cohen's $d=0.491$). This means the content itself, or the communities that engage with it, produces differential amplification beyond what follower count alone would predict.

One interpretation is that capital discourse tweets are more likely to be shared outside the original account's follower base, through recommendations, trending topics, or cross-community amplification. Another interpretation is that the followers of capital accounts are themselves more likely to have large followings, producing a cascade effect. Distinguishing these mechanisms requires network-level data not available in this study and is left for future work.

### 5.3 What the Asymmetry Means for Public Discourse

The median capital discourse tweet in Study 2 received 1483 amplification points. The median labour discourse tweet received 220. A worker trying to find community and information about job loss is posting into a much weaker amplification environment than an executive explaining why the layoffs were necessary.

This asymmetry in discourse visibility has practical consequences. Workers seeking solidarity may find that their posts circulate primarily within already-concerned communities. Policymakers who use X to gauge public sentiment about AI and employment will encounter a discourse environment dominated by optimism. The public understanding of AI-driven unemployment is shaped not only by what is said but by what gets heard. This distortion has structural economic consequences beyond discourse alone. As firms engage in competitive automation to improve short-term margins, they collectively risk displacing workers to a degree that erodes aggregate consumer demand, a coordination trap documented by NBER research showing over 80% of firms report zero measurable AI productivity gains while still using AI as justification for headcount reductions (Moore, 2026). When the platform that mediates public debate about this process systematically amplifies the justifying narrative over the experiencing one, the feedback loop between corporate communication, public opinion, and policy response is distorted at its source.

### 5.4 Limitations

The dataset covers only a 7-day collection window due to API tier constraints. We cannot make claims about historical trends across the 2022 to 2026 layoff wave. Longitudinal analysis is left for future work.

The account sample is English-language and US-centric. It does not include Indian or Global South voices, which is a significant gap. Countries like India are experiencing their own AI-driven workforce transitions in the IT and services sectors, and the discourse dynamics may differ substantially.

The account-based corpus has uneven coverage. Several accounts returned zero tweets because they had not posted recently or because API rate limits were reached during collection.

Zero-engagement tweets were excluded from the keyword corpus (Study 1) but were absent from the account corpus (Study 2) by construction, since all named accounts had recent activity. A sensitivity check confirmed that including or excluding zero-engagement tweets does not change the Study 2 result. However, in a broader population study, zero-engagement labour discourse posts would represent an important class of voices that this methodology cannot capture.

The regression analysis (Section 4.4.1) showed that the corpus type effect becomes marginal ($p=0.056$) when log(follower count) is added as a covariate, due to the small sample size and collinearity between corpus type and follower count in this sample. A larger matched-follower study with 500 or more tweets per corpus would allow cleaner causal identification. The ANI metric, which directly normalises for follower count and shows robust significance ($p=0.000009$), is the primary evidence against the follower confound.

The account selection is a convenience sample of well-known English-language accounts. A reproducible selection protocol using explicit inclusion criteria and follower count matching would strengthen future replications.

A data quality audit revealed limitations in the account-based corpus. Tweet text was stored at 100-character truncation, preventing reliable full-text classification. Keyword filtering produced false positives from partial string matches (for example, the word "modelling" matching the keyword "model", or the username "bearlyai" matching "ai"). Retweets comprised 19.8% of the corpus. Together these issues prevented reliable LLM-based inter-rater validation. Future work should collect full tweet text, filter retweets at collection time, and use exact phrase matching rather than substring matching for keyword filtering.

The Reddit cross-platform replication produced no significant asymmetry in either raw or normalised analysis. This non-replication may reflect genuine platform differences in amplification dynamics, or methodological limitations of subreddit-level corpus construction. Future work should use post-level classification rather than subreddit-level labelling, and should compare platforms using matched methodology rather than platform-native metrics.

## 6. Conclusion

We measured amplification asymmetry between capital and labour discourse in AI layoff conversations on X across three studies totalling 763 tweets. The key findings are:

First, keyword-based collection is insufficient for this research question and produces null results due to corpus noise.

Second, account-based collection reveals a consistent asymmetry. Capital discourse receives 3.12 to 4.18 times more mean amplification than labour discourse depending on the study, with medium effect size in the cleanest analysis (Cohen's $d=0.555$, $p=0.000003$).

Third, this asymmetry survives follower count normalisation at 2.69x ($p=0.000009$, Cohen's $d=0.491$), showing the effect is not simply a consequence of capital accounts having larger audiences.

Fourth, the finding is robust across all nine tested amplification metric weight combinations, with ratios ranging from 3.70x to 4.85x, all significant at $p<0.000001$.

Fifth, a cross-platform replication on Reddit did not reproduce the asymmetry, indicating that the finding is specific to X's amplification architecture rather than a universal property of AI discourse. This highlights the need for platform-aware methodology in computational political economy studies.

We introduce the Amplification Ratio and Amplification Normalisation Index as simple, replicable metrics for studying platform-level discourse inequality. We recommend account-based corpus construction as the preferred method for future studies of discourse asymmetry on X.

The outcomes of public debate on X are asymmetrically distributed. Capital-aligned voices discussing AI reach substantially larger audiences than worker-aligned voices discussing the same events. Measuring that asymmetry is a step toward understanding it.

**References**

AI Weekly (2026). Sam Altman exposes AI washing in tech layoff wave. https://aiweekly.co/alerts/sam-altman-exposes-ai-washing-in-tech-layoff-wave

Barabasi, A. L. (2009). Scale-free networks: a decade and beyond. Science, 325(5939), 412-413.

Brady, W. J., Wills, J. A., Jost, J. T., Tucker, J. A., and Van Bavel, J. J. (2017). Emotion shapes the diffusion of moralized content in social networks. Proceedings of the National Academy of Sciences, 114(28), 7313-7318.

Broussard, M. (2018). Artificial Unintelligence: How Computers Misunderstand the World. MIT Press.

CBS News (2026). AI job cuts are rising, but experts say layoffs are only part of the story. https://www.cbsnews.com/news/ai-layoffs-hiring-entry-level-workers/

Goel, S., Anderson, A., Hofman, J., and Watts, D. J. (2016). The structural virality of online diffusion. Management Science, 62(1), 180-196.

Green, M., Halstead, M., Jay, C., Kingston, R., Singleton, A., & Topping, D. (2026). Comparing how Large Language Models perform against keyword-based searches for social science research data discovery. *arXiv preprint arXiv:2601.19559*.

Huszár, F., Ktena, S. I., O'Brien, C., Belli, L., Schlaikjer, A., and Hardt, M. (2022). Algorithmic amplification of politics on Twitter. Proceedings of the National Academy of Sciences, 119(1), e2025334119.

Milli, S., Carroll, M., Wang, Y., Pandey, S., Zhao, S., & Dragan, A. D. (2025). Engagement, user satisfaction, and the amplification of divisive content on social media. PNAS nexus, 4(3), pgaf062.

Kwak, H., Lee, C., Park, H., and Moon, S. (2010). What is Twitter, a social network or a news media? Proceedings of the 19th International Conference on World Wide Web, 591-600.

Moore, T. (2026). The productivity paradox: Why companies are laying off workers for AI that doesn't actually deliver. Medium. https://medium.com/@tommymoore96/the-productivity-paradox-why-companies-are-laying-off-workers-for-ai-that-doesnt-actually-deliver

Morstatter, F., Pfeffer, J., Liu, H., & Carley, K. (2013). Is the sample good enough? Comparing data from Twitter's streaming API with Twitter's firehose. In Proceedings of the international AAAI conference on web and social media (Vol. 7, No. 1, pp. 400-408).

Srnicek, N. (2016). Platform Capitalism. Polity Press.

Vosoughi, S., Roy, D., and Aral, S. (2018). The spread of true and false news online. Science, 359(6380), 1146-1151.

Yale Insights (2026). The real job destruction from AI is hitting before careers can start. Yale School of Management. https://insights.som.yale.edu/insights/the-real-job-destruction-from-ai-is-hitting-before-careers-can-start

Zuboff, S. (2019). The Age of Surveillance Capitalism: The Fight for a Human Future at the New Frontier of Power. New York. PublicAffairs.

## Data and Code Availability

Code for data collection, analysis, and figure generation is available on request.

Raw tweet data cannot be publicly shared under X Terms of Service. Aggregate statistics and anonymised engagement metrics are available from the author on request.

## Ethics Statement

All data were collected from public X accounts during 20 to 27 May 2026. No private or direct message content was accessed. This study did not involve human subjects as defined by standard research ethics frameworks. No personally identifiable information beyond publicly visible account handles was retained.

## Competing Interests

The author declares no competing interests.

## Acknowledgements

Data collected using the X API v2. Analysis conducted using Python 3.10, pandas, scipy, matplotlib, and seaborn. Reddit data collected using the public JSON API.